\RequirePackage{amsmath} % defined before \documentclass to get rid of the error "Package amsmath Warning: Unable to redefine math accent \vec."
\documentclass[runningheads]{llncs}

\usepackage{amssymb}
\usepackage[pdftex,final]{graphicx}
\usepackage{marvosym} % for letter in author list

\usepackage[caption=false]{subfig}
\usepackage{interval}
\usepackage{siunitx}
\usepackage{multirow}
\usepackage{booktabs}

\graphicspath{{img/}}

\begin{document}

\title{Magnetic Resonance Fingerprinting Reconstruction via Spatiotemporal Convolutional Neural Networks}
\titlerunning{MRF Reconstruction via Spatiotemporal CNNs}
%\titlerunning{Abbreviated paper title}
% If the paper title is too long for the running head, you can set
% an abbreviated paper title here

%\author{$\ast \ast \ast$}
%\institute{$\ast \ast \ast \ast \ast$}

\author{Fabian Balsiger\inst{1}\textsuperscript{(\Letter)}\orcidID{0000-0001-7577-9870} \and Amaresha Shridhar Konar\inst{2} \and Shivaprasad Chikop\inst{2} \and Vimal Chandran\inst{1} \and Olivier Scheidegger\inst{3} \and \\ Sairam Geethanath\inst{2} \and Mauricio Reyes\inst{1}}

\authorrunning{F. Balsiger et al.} 
% First names are abbreviated in the running head.
% If there are more than two authors, 'et al.' is used.
%

\institute{Institute for Surgical Technology and Biomechanics,\\University of Bern, Bern, Switzerland
 %,\\ WWW home page:
%\texttt{http://www.istb.unibe.ch}
\and
Medical Imaging Research Center, Dayananda Sagar Institutions, Bangalore, India
\and
Support Center for Advanced Neuroimaging (SCAN), Institute for Diagnostic and Interventional Neuroradiology, Inselspital, Bern University Hospital, \\University of Bern, Bern, Switzerland \\
\email{fabian.balsiger@istb.unibe.ch}
}

\maketitle              % typeset the title of the contribution

% at least 70, maximum 150 words
\begin{abstract}

Magnetic resonance fingerprinting (MRF) quantifies multiple nuclear magnetic resonance parameters in a single and fast acquisition. Standard MRF reconstructs parametric maps using dictionary matching, which lacks scalability due to computational inefficiency. We propose to perform MRF map reconstruction using a spatiotemporal convolutional neural network, which exploits the relationship between neighboring MRF signal evolutions to replace the dictionary matching. We evaluate our method on multiparametric brain scans and compare it to three recent MRF reconstruction approaches. Our method achieves state-of-the-art reconstruction accuracy and yields qualitatively more appealing maps compared to other reconstruction methods. In addition, the reconstruction time is significantly reduced compared to a dictionary-based approach.

\end{abstract}

\section{Introduction}
Magnetic resonance imaging (MRI) is widely used in healthcare centers for the diagnosis of pathologies. The diagnosis from MRI relies mostly on weighted images, where the contrast between tissues is used to identify pathologies rather than the absolute intensities in the images. This qualitative approach limits the objective evaluation and reproducibility of MRI in the clinics. Although significant effort has been made for quantitative MRI, a clinical relevant solution for nuclear magnetic resonance (NMR) parameter mapping has not been achieved so far. Mainly time-inefficiency and the limitation to one NMR parameter at interest (e.g. T1 and T2 relaxation times) make quantitative MRI inappropriate for clinical use. To overcome the drawbacks of quantitative MRI, magnetic resonance fingerprinting (MRF) has been proposed recently as a novel quantitative MRI technique~\cite{Ma2013}. MRF quantifies multiple NMR parameters in a single and fast acquisition. The acquisition relies on a MR sequence with pseudo-randomly varying parameters to obtain a unique signal evolution, i.e. fingerprint, per tissue and voxel. After the acquisition, a dictionary matching algorithm assigns the voxel's signal evolutions to an entry of a dictionary of simulated and pre-computed signal evolutions, which allows reconstructing quantitative maps of NMR parameters of interest. However, this dictionary matching is time-consuming, lacks scalability, and can introduce artefacts due to the under-sampled \textit{k}-space during the acquisition~\cite{Wang2014}.

Recently, three approaches have been proposed aiming to overcome the issues associated with dictionary matching during the MRF reconstruction. G\'{o}mez et al.~\cite{Gomez2016} proposed a spatiotemporal dictionary matching that matches a spatial neighborhood of fingerprints instead of using a fingerprint-wise approach. They additionally improve the computational efficiency by limiting the matching to a local search window. However, the search window comes at the cost of requiring spatially aligned MRF scans, and ultimately only alleviates the problem of scalability of dictionary-based MRF reconstruction methods. Therefore, approaches replacing the dictionary matching using deep learning have been proposed to overcome the bottleneck of scalability. Cohen et al.~\cite{Cohen2018} proposed a fully-connected neural network and Hoppe et al.~\cite{Hoppe2017} proposed a convolutional neural network (CNN) to learn the matching of a MRF signal evolution to NMR properties. Both approaches show promising results regarding reconstruction accuracy and speed, and their concepts might be a feasible way to replace the dictionary matching involved in MRF reconstruction. However, they use a fingerprint-wise approach, i.e. do not consider any spatial characteristics during the reconstruction, which might result in noisy reconstructions. Moreover, all three approaches use maps reconstructed by the standard dictionary matching with simulated entries as ground truth to compare their reconstructed maps. This ultimately adds a bias to the methods, which resemble the dictionary matching instead of learning the underlying relation of the fingerprints to the NMR parameter maps.

We propose a MRF reconstruction approach that exploits the spatiotemporal relationship between neighboring signal evolutions motivated by noisy reconstructions of fingerprint-wise approaches and the findings of~\cite{Gomez2016}. Our approach bases on CNNs and yields fast and more accurate reconstructions than recently proposed methods on six healthy brain MRF images with three NMR maps: proton density (PD), T1 relaxation time (T1), and T2 relaxation time (T2). Unlike previously published methods, we rely on parametric maps acquired trough MR parameter mapping as ground truth instead of reconstructed maps by dictionary matching. We compare our performance to the aforementioned spatiotemporal dictionary- and deep learning-based methods. We report quantitative and qualitative results and discuss open issues and challenges towards a relevant solution for accurate and fast MRF reconstruction.

\section{Materials and Methods}
We consider a four-dimensional (4-D) MRF image $I \in \mathbb{C}^{X \times Y \times Z \times T} $, where each voxel $I(\mathbf{v}) = \{t_1, t_2, \dots, t_T\}$ at location $\mathbf{v} = (x, y, z)$ contains a MRF signal evolution, or fingerprint, with $T$ temporal signal intensities $t_i$. For each MRF image $I$, a set $Q = \{q_1, q_2, \dots, q_M\} \in \mathbb{R}^{X \times Y \times Z \times M}$ with $M$ parametric maps are available as ground truth for the reconstruction. In this work, six brain MRF images with $M = 3$ parametric maps $Q = \{\textnormal{PD}, \textnormal{T1}, \textnormal{T2}\}$ were used.

\subsection{MRF and Parametric Map Acquisition}
We acquired brain scans from six healthy male volunteers (21 to 43 years) using a tailored MRF sequence~\cite{Shaik2018} on a 1.5~Tesla GE SIGNA Artist scanner (GE Medical Systems, Milwaukee, WI, U.S.) with a 16-channel head coil as part of an institution approved study. Each scan consisted of $Z = 16$ axial-oriented slices with a matrix size of $X \times Y = 256 \times 256$, field of view (FOV) of $256 \times 256~\textnormal{mm}^2$, voxel size of $1.0 \times 1.0 \times 5.0~\textnormal{mm}^3$, and a total of 720 temporal images per slice. After the acquisition, the images were pre-processed using a sliding-window reconstruction~\cite{Cao2017} with a window size of 48 resulting in $T=673$ temporal images. 

% In tailored MRF, the flip angles (FA) are chosen to optimize the contrast for PD, T1 and T2 relaxation times in three independent blocks, which are combined into a single MRF sequence. The repetition times (TR) followed a Perlin noise pattern~\cite{Ma2013} while the FAs were generated based on 
% \begin{equation}
%   FA_t = \sin(2 \pi \frac{t}{500}) FA_{min},
% \end{equation}
% where $t$ is the TR index and $FA_{min} = \ang{5}$, $FA_{min} = \alpha_E$, and $FA_{min} = 2\alpha_E$ for the PD, T1, and T2 block, respectively. $\alpha_E$ is the Ernst angle. The acquisition is preceded by an inversion pulse (inversion time of 2750~ms) to suppress the signal from cerebrospinal fluid (CSF), and a delay of 5~s between the blocks was provided to decrease the effect of magnetization. The readout was a spiral trajectory that consisted of 48 arms with a readout time of 5~ms and a fixed echo time (TE) of 2.7~ms, which results in a total scan time per slice of 49~s.

The parametric maps serving as ground truth for the MRF reconstruction were acquired with the same number of slices, matrix size, FOV, and voxel size. The T1 and T2 maps were generated using curve fitting of the MR signal of multi-FA and multi-echo sequences, respectively. Seven T1-weighted images were acquired with a gradient recalled echo pulse sequence with FAs of \ang{1}, \ang{2}, \ang{5}, \ang{8}, \ang{11}, \ang{14}, and \ang{25}, and constant TR/TE = 5.85/1.77~ms. A fast spin echo sequence with eight TEs starting from 20~ms at an interval of 20~ms was used to generate the T2 map (FA = \ang{90} and TR = 1626~ms). By using a signal intensity equation, the PD maps were generated from the T1-weighted images acquired for the T1 mapping. 
%The B0 maps were generated through a dual echo GRE sequence with an echo separation of 1~ms for the off-resonance maps. 

\subsection{Spatiotemporal CNN MRF Reconstruction}
\begin{figure}[ht]
	\includegraphics[width=1\textwidth]{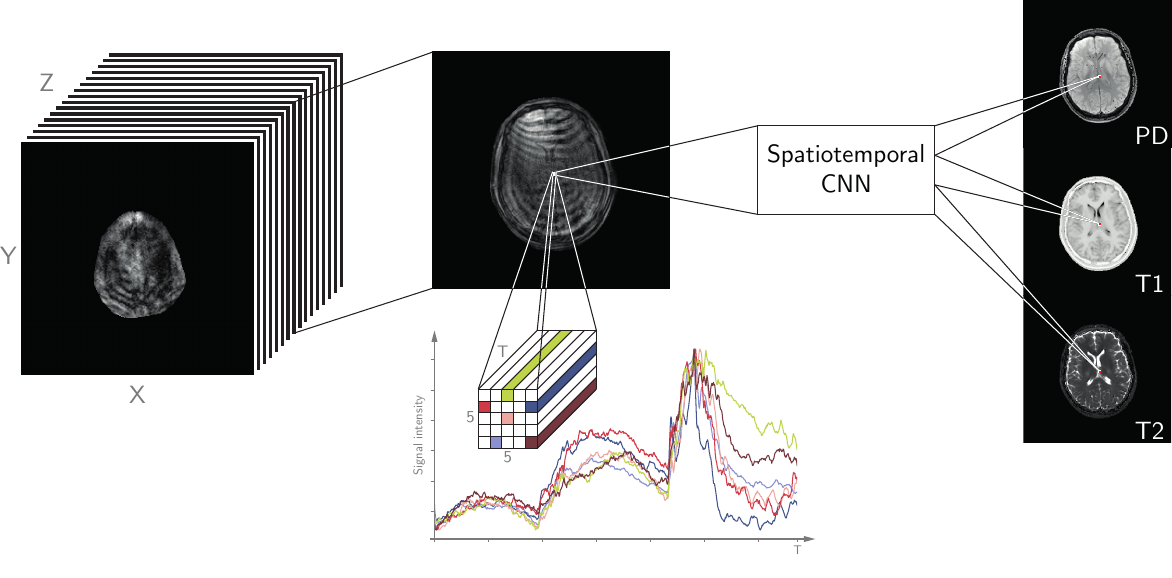}
	\caption{Overview of the proposed spatiotemporal MRF reconstruction. Note that the signal evolutions are complex-valued but the absolute numbers are plotted for simplicity.}
	\label{fig:overview}
\end{figure}

We propose a CNN to learn spatiotemporal features to reconstruct the maps $Q$ from a MRF image $I$. Input to the CNN are MRF image patches $I_P(\mathbf{v})\in \mathbb{C}^{5 \times 5 \times T} \subset I$, centered at location $\mathbf{v}$. Output of the CNN are the values of the estimated maps $\hat{Q}(\mathbf{v}) \in \mathbb{R}^{M}$ at location $\mathbf{v}$. The CNN is trained to learn the mapping $\mathcal{M}: I_P(\mathbf{v})\rightarrow Q(\mathbf{v})$. We remark that the reconstruction was performed slice-wise due to the large slice spacing of $5.0~\textnormal{mm}$ of our data. Figure~\ref{fig:overview} provides an overview of how the input and output data are defined for the proposed multiparametric spatiotemporal MRF reconstruction.

\subsubsection{Pre-processing}
We first apply a brain mask to the MRF images and the corresponding maps to exclude the background in all experiments. The masks were manually segmented using the T1 map with the polygon tool in ITK-SNAP (www.itksnap.org). Outliers from each map in $Q$ are removed by clipping the values to the percentiles $\interval{0.1}{99.9}$. Finally, we normalize $I$ along the temporal axis $T$ to have zero mean and unit variance, and $Q$ along the temporal axis $M$ to the range $\interval{0}{1}$. Note that within each subject the maps were spatially aligned and therefore no registration was applied.

\subsubsection{Architecture}
\begin{figure}[ht]
\centering
	\includegraphics[width=0.7\textwidth]{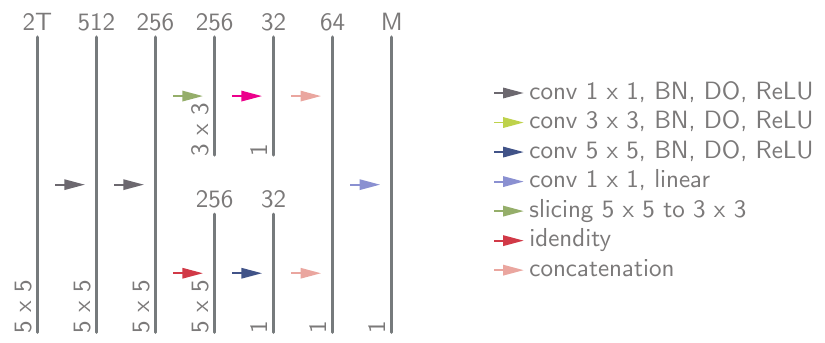}
	\caption{The architecture of our spatiotemporal CNN. To perform a convolution with a filter size of $3 \times 3$, we extract a $3 \times 3$ patch from a $5 \times 5$ patch denoted as slicing operation. The number of channels are denoted on the top of the bars and the $x \times y$ size is provided at the lower left edge of the bars. BN: batch normalization, DO: dropout.}
	\label{fig:architecture}
\end{figure}

Our network consists of five convolutional layers, which learn the mapping $\mathcal{M}$, i.e. we predict the $M$ map values at location $\mathbf{v}$ from an MRF image patch $I_P$ (Figure~\ref{fig:architecture}). We first concatenate the real and imaginary part of the complex-valued input $I_P(\mathbf{v})\in \mathbb{C}^{5 \times 5 \times T}$ to a real-valued input $I_P(\mathbf{v})\in \mathbb{R}^{5 \times 5 \times 2T}$ and consider the temporal dimension ($2T$) as the channels in our network. Second, we apply two $1 \times 1$ convolutional blocks to reduce the number of channels to 256. Subsequently, we apply two convolutional blocks in parallel with different receptive fields of $5 \times 5$ and $3 \times 3$ motivated by~\cite{Chen2016}. The output channels of these two convolutional blocks are concatenated and fed into the last convolutional layer with $M$ output channels corresponding to the values of the estimated maps $\hat{Q}$. A convolutional block in our network consists of a sequence of 2-D convolutional layer (valid padding and stride one), dropout, batch normalization, and rectified linear unit (ReLU) activation function. We maintain a linear activation at the last convolutional layer (valid padding and stride one). The estimated maps are denormalized to get quantitative values in the range prior to the pre-processing. We implemented our network using the open source machine learning library TensorFlow 1.8.0 (Google, Mountain View, CA, U.S.) with Python 3.6 (Python Software Foundation, Wilmington, DE, U.S.).

\subsubsection{Training} The network was trained using an Adam optimizer with a learning rate of 0.001, which minimized a mean squared error (MSE) loss with a batch size of 600 randomly sampled patches $I_P$. The dropout rate was set to 0.2 and the training was stopped after 50 epochs, which we empirically found to be sufficient. 
% ~\cite{Kingma2015}
\subsection{Evaluation}
We evaluate the performance our model and the baselines using a leave-one-out cross-validation, i.e. we train the model on five brain scans and test it on the left-out brain scan. Note that we tuned the architecture on one randomly chosen cross-validation split and did not use the other splits to develop and tune the architecture. 

\subsubsection{Baselines} We compare our method to recent approaches for MRF reconstruction using the fully-connected neural network~\cite{Cohen2018}, the CNN~\cite{Hoppe2017}, and the spatiotemporal dictionary matching~\cite{Gomez2016}. For the deep learning-based methods, we performed the same leave-one-out cross-validation and the data underwent the same pre-processing as for our method. The approaches were implemented as proposed in the papers. For~\cite{Gomez2016}, we also perform a leave-one-out cross-validation, i.e. construct a dictionary using five brain scans and reconstruct the left-out brain scan with following parameters: $W_n = 11 \times 11 \times 3$, $P = 3 \times 3 \times 3$, $C = 5$, and $\alpha = 0.5$ with two iterations.

\subsubsection{Metrics} Quantitatively, we report the mean and standard deviation of the mean absolute difference (MAE) and the root mean square error (RMSE) for the leave-one-out cross-validation. The metrics are reported separately for the three brain tissues white matter (WM), gray matter (GM), and cerebrospinal fluid (CSF). The brain tissue masks were obtained from the T1 maps using thresholding according to literature values~\cite{Ma2013}.

\section{Results}
\begin{figure}[t]
	\includegraphics[width=1\textwidth]{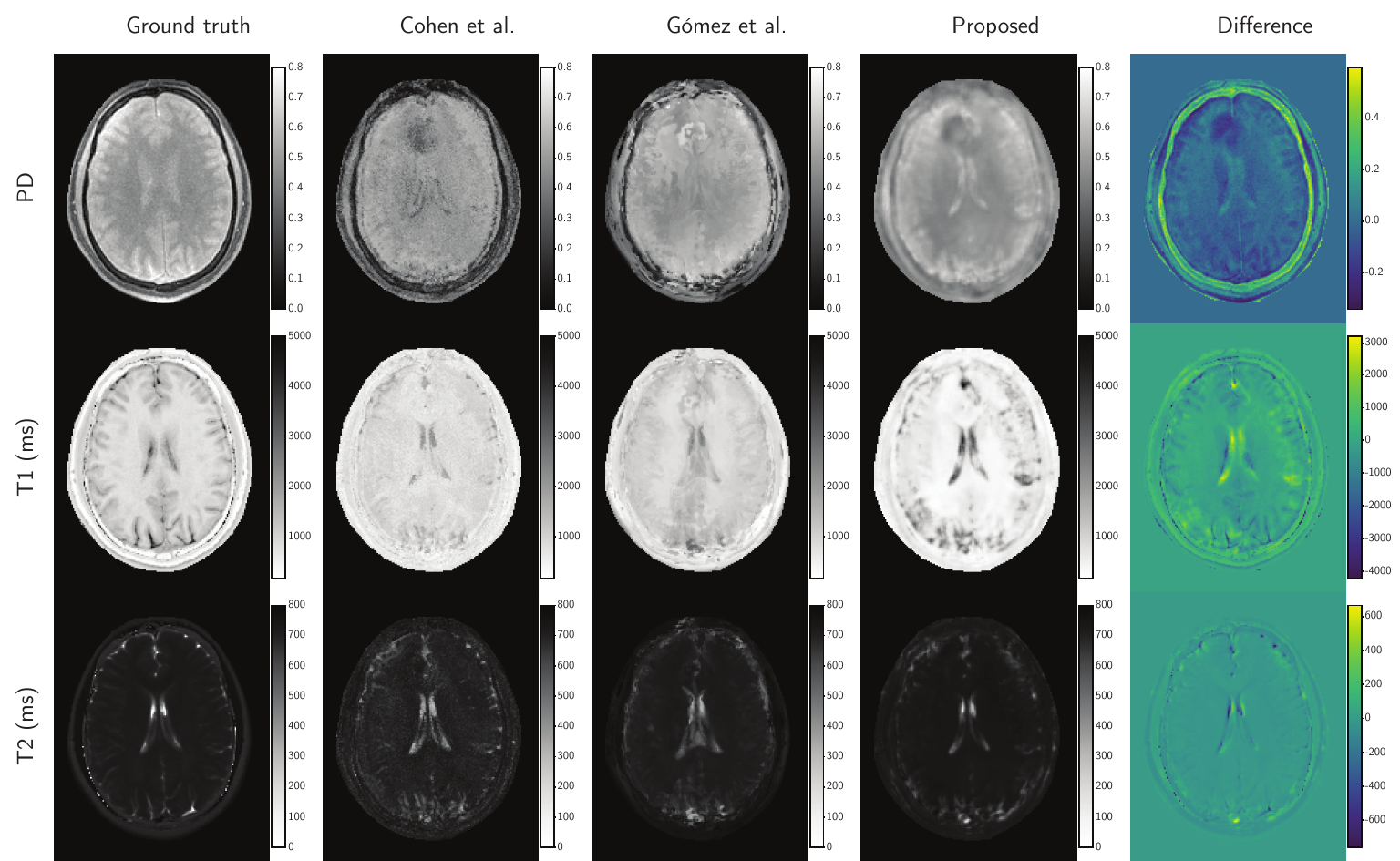}
	\caption{Exemplary map reconstructions of one axial brain slice. The rows represent the three maps PD, T1, and T2. The columns represent from left to right: the ground truth map, results of Cohen et al.~\cite{Cohen2018}, G\'{o}mez et al.~\cite{Gomez2016}, and our proposed method. The rightmost column shows the difference $\hat{Q} - Q$ between our estimated map $\hat{Q}$ and the ground truth map $Q$.}
	\label{fig:result1}
\end{figure}

Mean and standard deviation of the MAE and RMSE for the PD, T1, and T2 map reconstructions are given in Table~\ref{tab:result1}. The proposed method outperforms the other methods for most brain tissues and maps. Reconstructed maps of a mid-brain slice are shown in Figure~\ref{fig:result1}. Qualitatively, our maps show a good reconstruction with visible brain structures like the ventricles. It is noticeable that our spatiotemporal approach yields a less noisy reconstruction than the fingerprint-wise approach of Cohen et al.~\cite{Cohen2018} (similar noisy reconstructions were obtained for Hoppe et al.~\cite{Hoppe2017} but not reported in Figure~\ref{fig:result1}). The dictionary-based approach~\cite{Gomez2016} yields a qualitatively coarser reconstruction than our method. Overall, large reconstruction errors are mainly present at the skull, meninges, ventricles as well as at the boundary of the brain mask (rightmost column in Figure~\ref{fig:result1}), which could be consistently observed for all methods.

\begin{table}[ht]
\centering
\caption{Mean absolute error (MAE) and root mean square error (RMSE) for the PD, T1, and T2 map reconstructions separated by the brain tissues white matter (WM), gray matter (GM), and cerebrospinal fluid (CSF).}
\label{tab:result1}
\tiny
\begin{tabular}{l@{\hskip 1em} l@{\hskip 1em} c c@{\hskip 1em} c c@{\hskip 1em} c c@{\hskip 1em}}
\toprule
\textbf{Tissue} & \textbf{Method} & \multicolumn{2}{c}{\textbf{PD}} & \multicolumn{2}{c}{\textbf{T1 (ms)}} & \multicolumn{2}{c}{\textbf{T2 (ms)}} \vspace{.3em}\\

& & MAE & RMSE & MAE & RMSE & MAE & RMSE \\
\midrule
WM & Cohen & 0.084$\pm$0.030 & 0.107$\pm$0.032 & 209.0$\pm$28.3 & 267.5$\pm$21.5 & 43.6$\pm$23.9 & 77.3$\pm$54.7 \vspace{.3em} \\
& Hoppe & 0.080$\pm$0.030 & 0.101$\pm$0.031 & 253.9$\pm$64.3 & 317.7$\pm$67.2 & 61.3$\pm$36.3 & 92.0$\pm$53.8 \vspace{.3em} \\
& G\'{o}mez & 0.058$\pm$0.015 & 0.074$\pm$0.021 & 258.6$\pm$61.0 & 327.2$\pm$68.5 & 33.4$\pm$20.4 & 73.4$\pm$50.6 \vspace{.3em} \\
& Proposed & \textbf{0.055$\pm$0.015} & \textbf{0.072$\pm$0.016} & \textbf{159.4$\pm$36.3} & \textbf{242.7$\pm$54.7} & \textbf{28.0$\pm$17.6} & \textbf{71.1$\pm$62.0} \vspace{.5em} \\
GM & Cohen & 0.094$\pm$0.028 & 0.121$\pm$0.026 & 197.0$\pm$28.4 & \textbf{258.0$\pm$42.0} & 57.6$\pm$35.4 & 97.3$\pm$65.9 \vspace{.3em} \\
& Hoppe & 0.092$\pm$0.030 & 0.119$\pm$0.029 & 218.5$\pm$37.6 & 287.5$\pm$42.5 & 70.2$\pm$41.0 & 105.1$\pm$61.7 \vspace{.3em} \\
& G\'{o}mez & \textbf{0.060$\pm$0.017} & 0.081$\pm$0.021 & \textbf{190.8$\pm$24.7} & 269.1$\pm$43.1 & 45.8$\pm$26.9 & \textbf{90.5$\pm$59.6} \vspace{.3em} \\
& Proposed & 0.061$\pm$0.017 & \textbf{0.077$\pm$0.020} & 208.2$\pm$34.1 & 286.6$\pm$46.5 & \textbf{43.2$\pm$31.2} & 93.6$\pm$76.3 \vspace{.5em} \\
CSF & Cohen & 0.126$\pm$0.024 & 0.152$\pm$0.025 &1162.6$\pm$256.2 & 1364.1$\pm$265.1 & 183.3$\pm$54.1 & 237.0$\pm$58.0 \vspace{.3em} \\\
& Hoppe & 0.128$\pm$0.013 & 0.156$\pm$0.014 & 1013.3$\pm$236.0 & 1219.4$\pm$251.5 & \textbf{174.7$\pm$49.3} & \textbf{227.8$\pm$55.5} \vspace{.3em} \\
& G\'{o}mez & 0.102$\pm$0.020 & 0.129$\pm$0.019 & 1072.5$\pm$164.5 & 1268.8$\pm$172.6 & 228.6$\pm$85.7 & 286.8$\pm$86.0 \vspace{.3em} \\
& Proposed & \textbf{0.093$\pm$0.013} & \textbf{0.113$\pm$0.009} & \textbf{989.2$\pm$254.7} & \textbf{1181.5$\pm$288.6} & 181.6$\pm$48.6 & 240.2$\pm$48.7 \vspace{.5em} \\
\bottomrule
\end{tabular}
\end{table}

\section{Discussion and Conclusion}

% CNN learn NN, Gomez kNN. Our results are worse but ultimate goal to compare to real parameters instead to the dictionary machting

We presented a deep learning-based, dictionary-free approach to reconstruct parametric maps from MRF images that exploits the spatiotemporal relationship between neighboring fingerprints. The approach is designed as CNN that yields a reconstruction of parametric maps in a more accurate way than previously proposed dictionary-free methods and competes with a dictionary-based method.

In general, the results show that a spatiotemporal reconstruction is favorable to a fingerprint-wise reconstruction for almost all brain tissues and parametric maps (Table~\ref{tab:result1}). Out of the three brain tissues, the GM yielded the most inconsistent results among the different methods. We think that this might arise due to partial volume effects at the interface to WM and CSF. A spatial analysis reveals high reconstruction errors in the skull, meningeal layers, and ventricles for all methods (rightmost column in Figure~\ref{fig:result1}). These reconstruction errors could origin from a partial volume effect or an apparent lack of training examples. Reconstruction artifacts are only present in the method of G\'{o}mez et al.~\cite{Gomez2016}, confirming the findings of~\cite{Wang2014}. In regards to the computational costs, all deep learning-based approaches yield reconstructed maps within few seconds. Conversely, the dictionary-based approach is computationally intensive, with calculations in the order of several minutes per reconstruction. For clinically used MRF reconstruction, we therefore think that machine learning-based approaches are favorable to dictionary-based approaches in the long term.

Our study design included parametric maps acquired trough MR parameter mapping as ground truth. The reconstruction errors of all methods are large compared to the errors reported in the studies of the baselines~\cite{Cohen2018,Gomez2016,Hoppe2017}. Such large errors are especially surprising for the dictionary-based method, which can be interpreted as a k-nearest neighbor search. Unfortunately, all baselines compared their performance with a ground truth obtained by dictionary matching as proposed in the original MRF paper~\cite{Ma2013}. Therefore, the methods resembled the dictionary matching instead of learning the underlying relation between fingerprints and NMR maps. We think that a comparison to acquired NMR maps is ultimately more meaningful than a comparison with maps reconstructed from simulated dictionaries. Our results suggest that the direct learning of fingerprints to acquired NMR maps is possible, although additional investigations and work are needed to lower the reconstruction errors.

In conclusion, we demonstrated that a spatiotemporal MRF reconstruction is favorable to a fingerprint-wise MRF reconstruction designed within a CNN by achieving quantitatively and qualitatively better parametric map reconstructions. 

\subsubsection{Acknowledgements}
This research was supported by the Swiss Foundation for Research on Muscle Diseases (ssem), grant attributed to author OS. The authors thank the NVIDIA Corporation for their GPU donation.

% ---- Bibliography ----
\bibliographystyle{splncs04}
\bibliography{library}

\end{document}